\documentclass{bmvc2k}
\usepackage[skip=2pt]{caption}
\usepackage{subcaption}
\usepackage{booktabs}
\usepackage{multirow}
\usepackage{amssymb,amsmath,amsthm,enumitem}

\title{Track Targets by Dense Spatio-Temporal Position Encoding}

\addauthor{Jinkun Cao}{jinkunc@andrew.cmu.edu}{1}
\addauthor{Hao Wu}{wuhao.5688@bytedance.com}{2}
\addauthor{Kris Kitani}{kkitani@cs.cmu.edu}{1}

\addinstitution{
Carnegie Mellon University
}
\addinstitution{
ByteDance
}

\usepackage{color, colortbl}
\usepackage{mathtools}
\usepackage{amsfonts}
\runninghead{J. CAO ET Al.}{Dense Spatio-Temporal Position Encoding}
\definecolor{aliceblue}{rgb}{0.94, 0.97, 1.0}
\definecolor{bananamania}{rgb}{0.98, 0.91, 0.71}

\begin{document}

\maketitle

\begin{abstract}
In this work, we propose a novel paradigm to encode the position of targets for target tracking in videos using transformers. The proposed paradigm, Dense Spatio-Temporal (DST) position encoding, encodes spatio-temporal position information in a pixel-wise dense fashion. The provided position encoding provides location information to associate targets across frames beyond appearance matching by comparing objects in two bounding boxes. Compared to the typical transformer positional encoding, our proposed encoding is applied to the 2D CNN features instead of the projected feature vectors to avoid losing positional information. Moreover, the designed DST encoding can represent the location of a single-frame object and the evolution of the location of the trajectory among frames uniformly. Integrated with the DST encoding, we build a transformer-based multi-object tracking model. The model takes a video clip as input and conducts the target association in the clip. It can also perform online inference by associating existing trajectories with objects from the new-coming frames. Experiments on video multi-object tracking (MOT) and multi-object tracking and segmentation (MOTS) datasets demonstrate the effectiveness of the proposed DST position encoding. 
\end{abstract}

\section{Introduction}
The transformer~\cite{vaswani2017attention,detr} has introduced a new powerful paradigm for processing sequential data. Among the innovations by transformers, positional encoding is an essential addition to the transformer. It provides information of token position for 1D text sequences. However, compared to its success in language models, positional encoding plays a relatively minor role in many vision tasks, such as multi-object tracking. When applying transformers in multi-object tracking, popular methods~\cite{transtrack,meinhardt2022trackformer,gtr} still mostly rely on appearance matching to associate targets across multiple time steps. 

Typically, the positional encoding is added to the tokens in the transformer to provide information about the relative order of the input tokens. It has properties such as being consistent for token pairs with the same relative distance, making it ideal for processing 1D sequences of text tokens. However, when using positional encodings in vision tasks, the previously defined position encoding is less well-formed to preserve position information in images (2D) and video tubes (3D). Consequently, many transformer-based methods have found positional encoding ineffective, especially in target tracking~\cite{gtr,transtrack} tasks, and have stuck to applying appearance similarity as the cue to associate targets.

However, we believe position information should play a more critical role in multi-object tracking. Moreover, by recognizing the flaws of directly migrating positional encoding from language processing to vision tasks, we find that the key is to keep spatial and temporal information lossless in the positional encoding. Motivated by such analysis, we propose a new paradigm of applying the positional encoding earlier, on 2D CNN feature maps, rather than later, on projected feature vectors. We could preserve pixel order and positional information now. 
Furthermore, we take advantage of the natural Fourier properties of our proposed positional encoding. By approximating the underlying Fourier and maintaining its linearity, we can achieve a uniform position encoding form for detections and trajectories. This enables the model to associate (1) among the detections and (2) between the detections and the trajectories in the same way. 
As the proposed positional encoding spans every pixel densely and can represent the pixel position evolution over time, we name it Dense Spatio-Temporal position encoding or DST encoding. We also propose using an attention mask for more accurate pixel-wise feature extraction and to avoid noise from background pixels. The attention mask can be computed from either segmentation masks, saliency discovery maps, or other coarse pixel-wise maps.

With the proposed DST encoding, we build a transformer-based method achieving state-of-the-art performance on multi-object tracking and multi-object tracking and segmentation benchmarks. We also provide an analysis of the shortcomings of classic positional encoding and how our DST encoding improves upon it as a new baseline for future works. 

\section{Related Works}
\subsection{Tasks for Tracking Targets in Videos}
Topics related to Target tracking in videos include multi-object tracking (MOT), multi-object tracking and segmentation (MOTS), video object/instance segmentation (VOS/VIS), and segmenting and tracking every pixel (STEP). We choose MOT~\cite{milan2016mot16,sun2022dancetrack} and MOTS~\cite{voigtlaender2019mots} to evaluate our proposed method because there are multiple targets in the video and they show long-range movement, making them suitable tasks to verify the effectiveness of our proposed method. 
On the contrary, VOS/VIS datasets, such as DAVIS~\cite{davis} and Youtube-VOS~\cite{xu2018youtube}, contain foreground objects of very different appearances or even categories and they usually have simple and slow movement. 
On the other hand, STEP~\cite{weber2021step} is based on MOTS but adds static objects to consider, such as buildings, road lanes, and trees. These objects are static and easy to track by linear motion models and are not suitable for showcasing the advantages of our DST position encoding.

\subsection{Positional Encoding as a Representation}
The currently widely used positional encoding is introduced by the transformer~\cite{vaswani2017attention} for language models and then extended to vision tasks~\cite{detr}.
Positional encoding or its variants with different names has been studied for a long time as a form of representation. An early work~\cite{rahimi2007random} has studied random Fourier features to approximate an arbitrary stationary kernel using Bochner's theorem. It is close to the use of positional encoding in the transformer. 
In computer vision, coordinate-MLPs provide a way to encode objects' positions as weights and are related to the study of positional encoding~\cite{zhong2019reconstructing,tancik2020fourier}. More recently, ~\citet{zheng2021rethinking} also suggest a study of positional encoding beyond a Fourier lens. They show that non-Fourier embeddings can also serve as positional encoding and, in the perspective of coordinate-MLPs, the performance is determined by a trade-off between embedding matrix stable rank and the distance preservation of coordinates. However, all these explorations have not suggested an efficient form of positional encoding for vision tasks to preserve the spatial transformation of a series of positions.

\subsection{Multi-objec Tracking Algorithms}
Early works on multi-object tracking mainly focus on motion analysis on the target trajectory, where the Kalman Filter is a classic solution~\cite{bewley2016simple}. 
Later, the rise of deep learning brings the powerful deep visual representations and related algorithms follow two paradigms: tracking-by-detection and joint-detection-and-tracking methods. Both of these paradigms involve an association stage, where they mostly focus on appearance matching~\cite{zhang2021fairmot,pang2021quasi}, i.e., re-identification, without using the motion information. 
More recently, transformer~\cite{vaswani2017attention} is introduced into the area of multi-object tracking~\cite{transtrack,meinhardt2022trackformer,motr} to take advantage of its parallel processing power. However, existing methods still neglect the information motion information, with the exception of  MOTR~\cite{motr} which has attempted to model motion implicitly using a query iteration mechanism. GTR~\cite{gtr} shows that using position encoding decreases transformer performance on MOT tasks. All the evidence suggests that the existing ways for leveraging motion and position information in transformer trackers are ineffective, which motivates the explorations in this paper.

\section{Method}
In this section, we first provide an overview of the architecture of our method and then detail its components: the design of the Dense Spatio-Temporal (DST) encoding, the attention mask, and the training and inference configurations. 

\begin{figure}
    \centering
    \includegraphics[width=.98\linewidth]{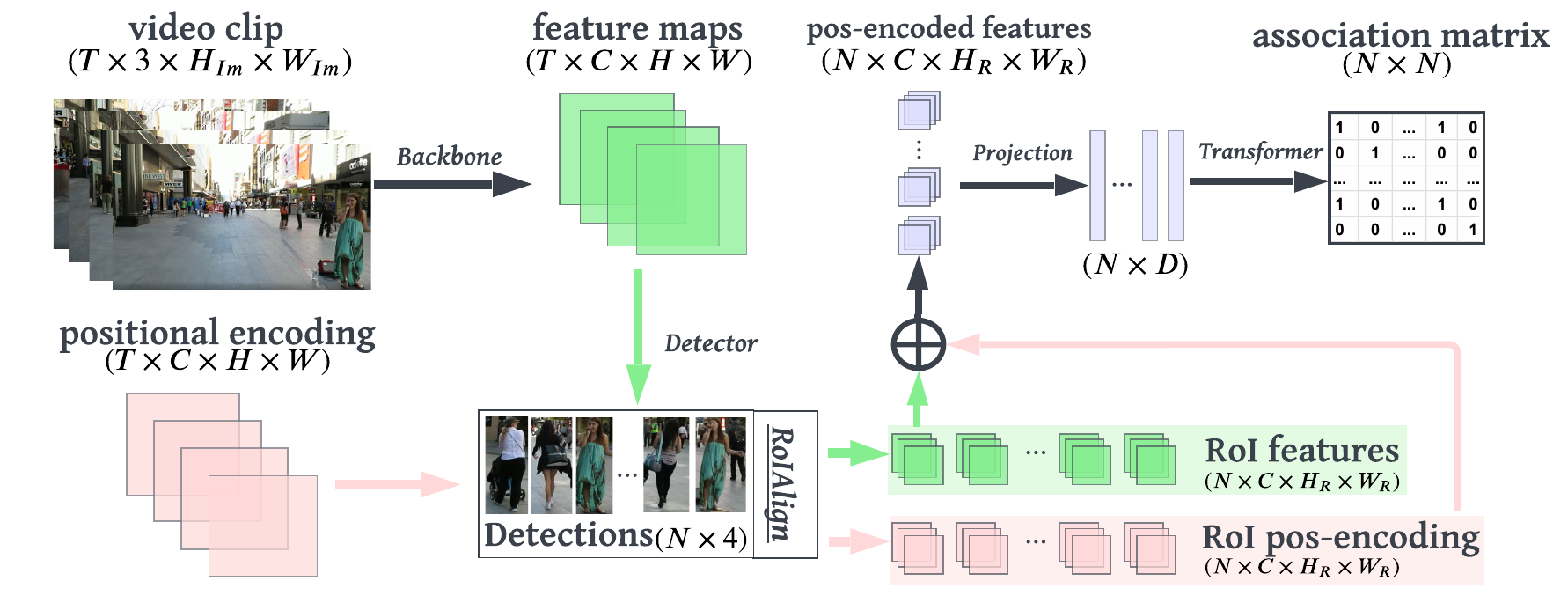}
    \caption{The illustration of associating targets within a video clip. For general cases, we use bounding boxes to represent the targets of interest while we can further replace the detector with a segmentation model to do the attention in a more fine-grained mask area. Positional encoding is added to CNN feature maps to encode position information.}
    \label{fig:arch}
    \vspace{-5mm}
\end{figure}

\subsection{Overview}
\label{sec:structure}
The proposed method can make associations at two levels: between detections in a video clip or between detections and existing trajectories.

\paragraph{Association of detections in a video clip.} 
For the association of objects, we follow the ``global association'' scheme widely adopted by transformer-based methods~\cite{gtr,motr,vistr},  as shown in Figure~\ref{fig:arch}. With the images of $T$ frames as input, we first use a backbone network to extract the feature maps. Then, a detector head is used to localize $N$ objects of interest inside these images with optional segmentation to gain more fine-grained feature representation. Given the localized objects, we extract their RoI features on both CNN features maps and DST encoding, which are of the same shape $T \times C \times H_R \times W_R$, 
where $(H_R, W_R)$ is the preset size of RoI, e.g. $7 \times 7$. Finally, we add both features and project them to feature embeddings of size $N \times D$, which we then forward into a transformer decoder to compute the attention score matrix of size $N \times N$. Considering that there should be no association between objects from the same frame, we perform softmax on each frame respectively to ensure a well-formed association matrix.

\begin{figure}
\centering
        \begin{subfigure}[b]{0.55\textwidth}
                \centering                \includegraphics[width=\linewidth]{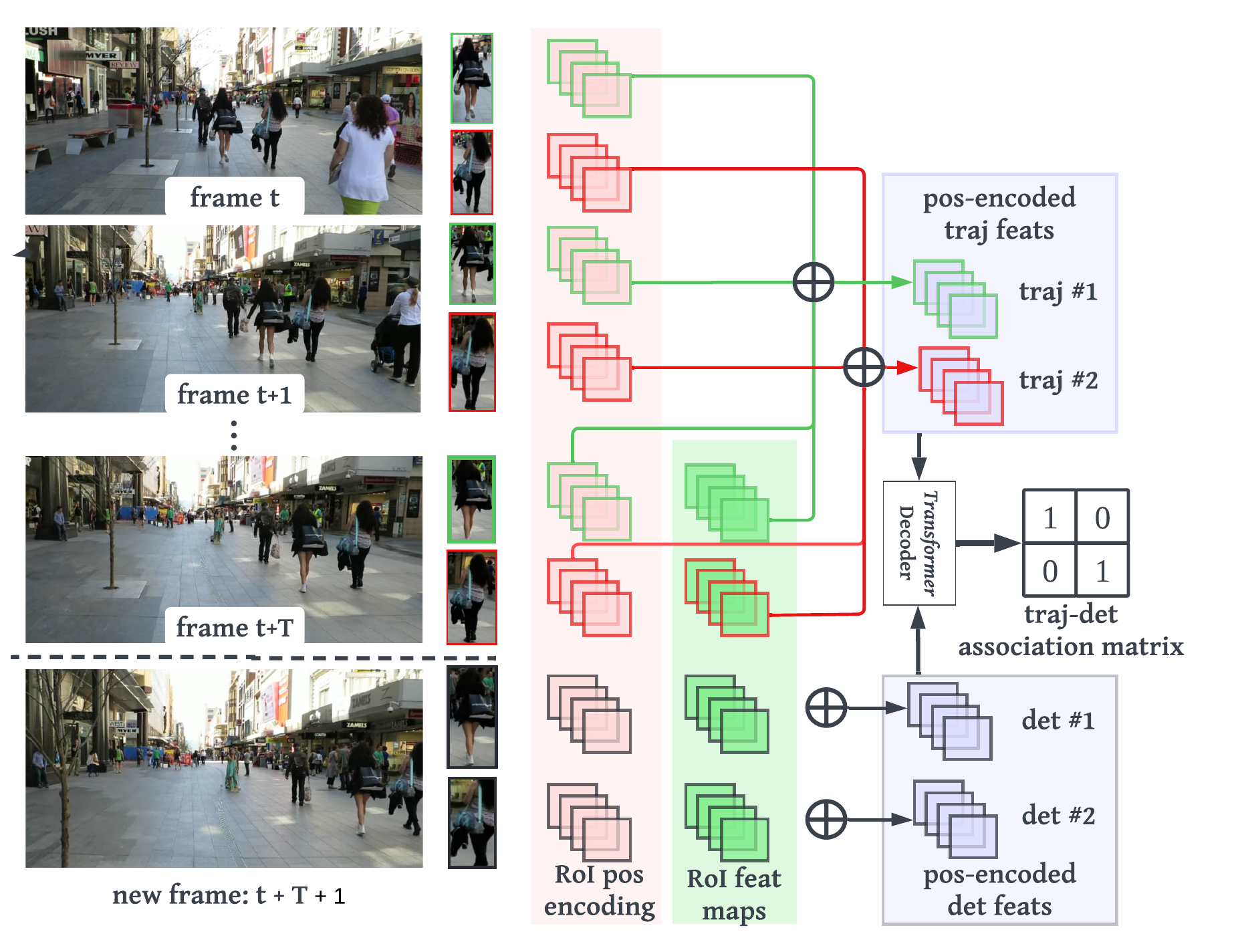}
                \caption{}
                \label{fig:det_traj}
        \end{subfigure}%
        \begin{subfigure}[b]{0.38\textwidth}
                \centering
             \includegraphics[width=\linewidth]{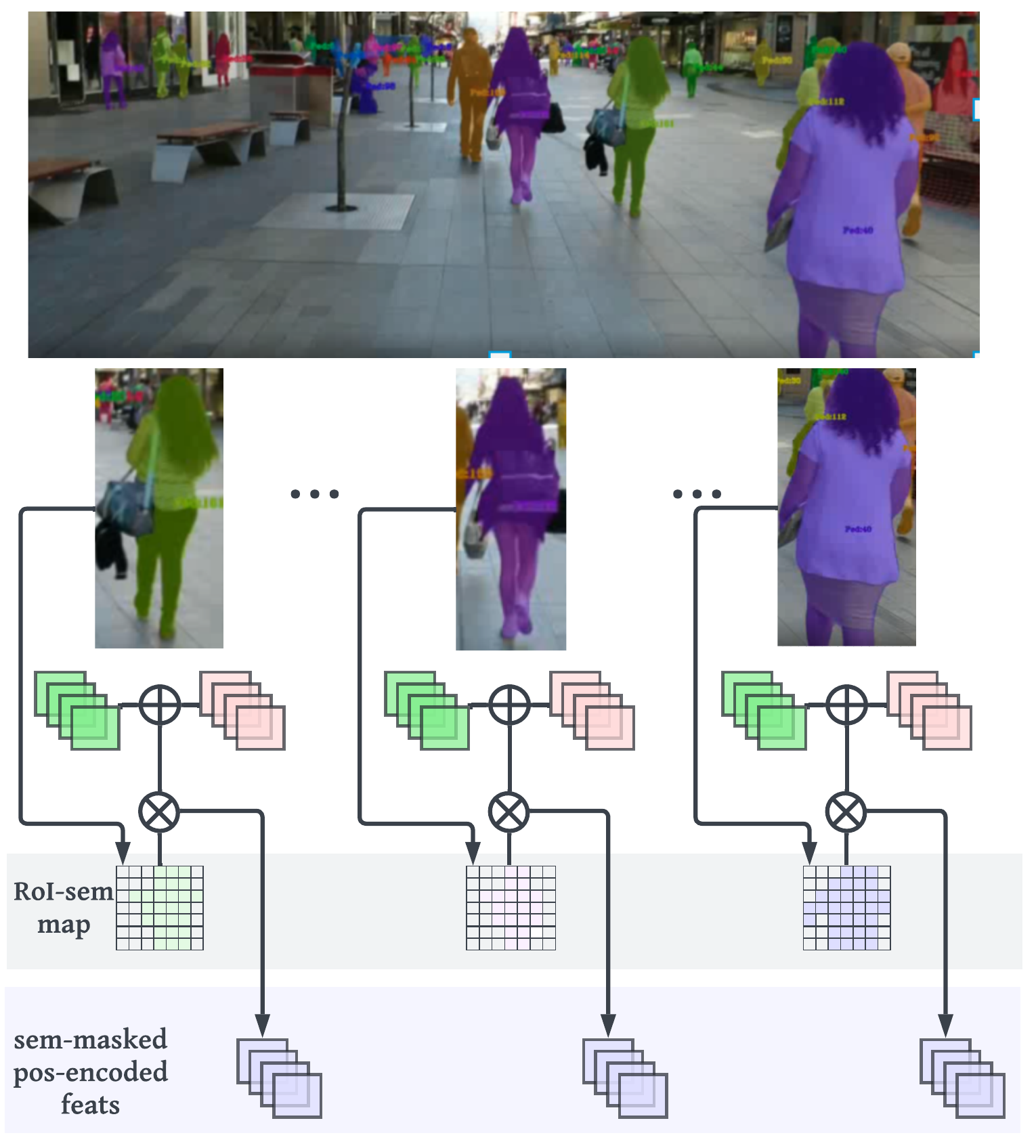}
                \caption{}
                \label{fig:roi_sem}
        \end{subfigure}%
        \caption{A deeper look at the component in our method. (a): how we generate the feature representations for both trajectories and single-frame objects. The representation of a trajectory is the accumulated positional encoding of all contained historical locations and the appearance feature of the last snapshot of the object. (b): for the video tracking and segmentation task, we use the semantic occupancy map onto object RoI to obtain more fine-grained RoI features where both position and semantics are encoded.}
        \label{fig:detailed}
\end{figure}

\paragraph{Association between detections and trajectories.} The proposed method can also perform the association between the detections on a new-coming video frame and existing trajectories for online tracking during inference. During the online inference, we perform tracking frame by frame by using a sliding window on the video with a stride of 1.
We align the representation of trajectories in the same shape and form as detections to enable this process to share the same model for the detection-detection association.
To represent the position of detections on a single frame, we apply RoI to extract the corresponding area from the DST map. However, to represent the trajectory, we now have to use the accumulated DST encoding to record the positional evolution of the track.
In this fashion, the representation of a trajectory is designed to be the element-wise addition of accumulated DST encoding of historical object positions and the CNN features of the object snapshot at the last frame.
The process of associating detections and trajectories is explained in Figure~\ref{fig:det_traj}. We perform softmax over the dimension of detections and the dimension of trajectories, respectively, to output the final association matrix. We use the Hungarians algorithm to ensure an one-to-one mapping between detections and trajectories. If a detection's attention score with all trajectories is lower than a threshold $\beta$ or all available trajectories are already associated, this remaining detection will give birth to a new trajectory.

\subsection{Dense Spatio-Temporal Position Encoding}
\label{sec:encoding}
Re-identification-based tracking methods associate targets across frames by comparing the appearance similarity of targets, neglecting the location information. However, we believe that location cues can significantly help associate targets because objects usually follow certain motion patterns in the real world.
To present the location of each object, a navie way is to append the bounding box coordinates to the object's feature vector. However, this operation can not scale up to trajectories of arbitrary length. Recently, the transformer has been adopted in multi-object tracking with the one-dimensional sinusoidal positional encoding~\cite{vaswani2017attention} added to token vectors. But it is not as effective~\cite{transtrack,motr,gtr} as in the language process tasks~\cite{vaswani2017attention}. We argue that to scale the 1D positional encoding up to 2D images or 3D videos, we need to avoid the loss of spatial information during feature projection. We will demonstrate that our proposed DST encoding can solve this problem to provide a better structure of location information in representing object trajectories.

\paragraph{Encoding of single-frame locations.} Given the channel number of feature maps is $C$, for a pixel at position $(x,y)$ in the image (or feature maps) whose size is $W \times H$, its positional encoding value at the $i$-th channel is
\begin{equation}
\small
    P(x, y, i) = 
    \begin{cases}
        -cos \left [ (\frac{x}{W} + \frac{y}{WH}) \pi + \frac{2i\pi}{C} \right ], i = 2k +1 \\
        cos \left [ (\frac{y}{H} + \frac{x}{WH}) \pi + \frac{2i\pi}{C} \right ],  i = 2k\\
    \end{cases}
   k \in \mathbb{Z} \cap [0,\frac{C}{2}).
\end{equation}

Such an encoding has a few desirable properties. First, it injectively maps from the pixel position to a value on all channels of the feature maps. Second, it keeps the encoding zero-centered spanning the image area which is friendly to the model training. Finally, the term $\frac{2i\pi}{C}$ keeps the encoding fairly sensitive to location variance within the whole area of the image. Without this term, the encoding value changes more sensitively around the image center while less sensitively near the image boundary. This is easy to prove by checking the first derivative of the sinusoidal function.

What we use in the final DST encoding is the resized encoding from the RoI area of objects only. This helps the model to have an encoding of the fixed shape and focus on the object area. If the shape of RoI is $W_R \times H_R$ and the bounding box coordinates of an object on the raw image is $(u, v, u+w, v+h)$, on the cropped and resized RoI feature maps, the positional encoding becomes 
\begin{equation}
\small
\begin{aligned}
    P_R(x^\prime, y^\prime, i) &= 
    \begin{cases}
        -cos \left [ (\frac{w}{WW_R}x^\prime + \frac{h}{WHH_R}y^\prime)\pi + (\frac{u}{W}  + \frac{v}{WH}) \pi + \frac{2i\pi}{C} \right ], i = 2k +1 \\
        cos \left [  (\frac{h}{HH_R}y^\prime + \frac{w}{WHW_R}x^\prime) \pi + (\frac{v}{H}  + \frac{u}{WH} ) \pi + \frac{2i\pi}{C} \right ],  i = 2k\\
    \end{cases} \\ 
\end{aligned} k \in \mathbb{Z} \cap [0,\frac{C}{2}),
\end{equation}
where $x^\prime \in [0, W_R]$ and $y^\prime \in [0, H_R]$, only now extending in the boundary box area. Here, the period of this encoding function changes in terms of the ratio of object size and RoI size. Therefore, this operation also implicitly encodes the target shapes instead of just the position.

\paragraph{Encoding of trajectory.}  On two time steps $t_1$ and $t_2$, we note the bounding boxes of a target object as $\mathbf{b}_1 = (u_1, v_1, u_1+w_1, v_1+h_1)$ and $\mathbf{b}_2 =  (u_2, v_2, u_2+w_2, v_2+h_2)$. Now, by adding the positional encoding in the RoI area, we have the trajectory encoding of every pixel in the two bounding boxes as 
\begin{equation}
\begin{aligned}
    P_R^{\mathbf{b}_2 | \mathbf{b}_1  }(x^\prime, y^\prime, i)  &= P_R^{\mathbf{b}_1} (x^\prime, y^\prime, i) + P_R^{\mathbf{b}_2} (x^\prime, y^\prime, i). \\ 
\end{aligned}
\end{equation}
Because the period of function $P_R^{\mathbf{b}_2 | \mathbf{b}_1}$ is still longer than $W_R$ and $H_R$ on the direction of width and height, it can still represent the trajectory from $\mathbf{b}_1$ to $\mathbf{b}_2$ injectively. Furthermore, we can extend this trajectory encoding to longer video clips as 
\begin{equation}
    P_R^{\mathbf{b}_T | ... | \mathbf{b}_1} (x^\prime, y^\prime, i) = \sum_{t=1}^T \alpha_t P_R^{\mathbf{b}_t} (x^\prime, y^\prime, i),
\end{equation}
where $\alpha_t$ is the weighting factor on the $t$-th frame. As for each frame, we have the dense position encoding on each pixel in the object area in the form of trigonometric functions; the trajectory encoding is well represented in a Fourier series now. We choose a linear combination of frame-wise encoding to take advantage of the linearity of Fourier series that is $\mathcal{F}(\sum_{i=1}^K \sigma_i f_i) = \sum_{i=1}^K \sigma_i \mathcal{F}(f_i)$,
where $\mathcal{F}$ is the Fourier transform and $\sigma_i$ is the weighting factor for function $f_i$. This property ensures the sanity to extend trajectory encoding by linearly adding the position encoding on the new coming frame. To show this, we note $\mathcal{T}^{\mathbf{b}_T|...|\mathbf{b}_1}$ the underlying function that we aim to approximate to represent a trajectory along the bounding boxes $(\mathbf{b}_1,...,\mathbf{b}_T)$. Then, if we have a function $\mathcal{L}$ that maintains the linearity, we have 
$\mathcal{F}(\mathcal{T}^{\mathbf{b}_T|...|\mathbf{b}_1}) = \mathcal{L}(P_R^{\mathbf{b}_T | ... | \mathbf{b}_1})$. Therefore, extending the trajectory to the next position $\mathbf{b}_{T+1}$  keeps the form of the positional encoding for the trajectory the same:
\begin{equation}
    \mathcal{L}(P_R^{\mathbf{b}_{T+1} |\mathbf{b}_T | ... | \mathbf{b}_1}) = \mathcal{L}(P_R^{\mathbf{b}_T | ... | \mathbf{b}_1}) + \mathcal{L}(P_R^{\mathbf{b}_{T+1}}) = \mathcal{F}(\mathcal{T}^{\mathbf{b}_T|...|\mathbf{b}_1}) +  \mathcal{F}(\mathcal{T}^{\mathbf{b}_{T+1}}) = \mathcal{F}(\mathcal{T}^{\mathbf{b}_{T+1}|\mathbf{b}_T|...|\mathbf{b}_1}).
\end{equation}

Now, we have shown that the proposed DST encoding can preserve the position information in a spatio-temporal occupancy tube densely and at arbitrary length. On each encoding channel, the value is variant to both the absolute position of the corresponding pixel and the position difference across frames. On the other hand, the traditional positional encoding in the transformer maintains the same encoding for any tokens of the same position difference. Also, since the full period ($2\pi$) spans on the feature channel dimension ($C$), it can always map the same relative position shift of two pixels to the same value on different channels. In practice, we use an MLP without non-linear activation to model the function $\mathcal{L}$ along the dimension of encoding channels.
If a target moves smoothly along the width and height directions, the encoding of its previous trajectory and its encoding on a new-coming frame will output a high similarity by attention.

Compared to the classic vector positional encoding, DST encoding has three main advantages: (1) preserving the object location information; (2) encoding pixel-wise dense information; (3) unifying representation for single-frame objects and trajectories across multiple frames. These properties provide additional knowledge to associate targets across frames.

\subsection{Dense Spatio-Temporal Attention}
\label{sec:details}
As both visual features and location encoding are dense on every pixel, we can do the association in a pixel-wise dense fashion now. But in fact, the target objects often change their pose in the bounding box and the bounding box includes background area as noise, especially when the targets are non-rigid such as the human body in pedestrian tracking. But when the video frame rate is high, the relative movement of the object body inside the bounding box is minor, dense attention is still very useful. Moreover, we perform attention to the RoI elements instead of the raw image pixels. Each pixel in RoI is already a conclusion of multiple pixels on the raw images. It makes dense attention more robust. For the association of detections in a video clip, the features already integrated with positional encodings are noted as $F \in \mathbb{R}^{N \times C \times H_R \times W_R}$ for $N$ objects. Then, we apply attention mask $M \in \mathbb{R}^{N \times H_R \times W_R}$ determining which ``pixel'' in the RoI areas should be attended to. 
In practice, the attention mask $M$ can be the segmentation mask (Figure~\ref{fig:roi_sem}) if that is available or an attention map without using segmentation supervision. We copy the feature along the channel dimension to scale it to $M^\prime \in \mathbb{R}^{N \times C \times H_R \times W_R}$. Next, we apply an MLP to transform the features into 1-d feature vectors, the operation noted as $g(\cdot)$. Given all the preparation, we get the encoded feature vector as $g(M^\prime F)$, which would later be transformed to $K$ and $Q$ by linear layers in self-attention. Finally, we predict the attention matrix as 
$S = \text{softmax}(\frac{Q \times K^T}{\sqrt{D}})$.
This also works in the case of cross-attention for associating trajectories and detections. For a trajectory, $M$ is the attention mask on its last frame. We will apply Hungarians algorithm to ensure the validity of the final binary association matrix from the attention matrix. 

\subsection{Training and Inference}
\paragraph{Training.} During training, we draw $N$  high-confidence detections from a detector after NMS, noted as $\mathcal{D} = \{D_1, ..., D_N\}$. The features with positional encoding added are noted as $\{F_1, ..., F_N\}$. From the self-attention-based association of objects within the video clip, we can output its association matrix as $\hat{S} \in \mathbb{R}^{N \times N}$. With the ground truth association matrix $S \in \mathbb{R}^{N \times N}$, we can derive the MSE loss for in-clip object association as
\begin{equation}
    l_{clip}(S, \hat{S}) = \frac{1}{N^2}\sum_{i,j}(S_{i,j} - \hat{S}_{i,j})^2.
\end{equation}
In addition to this, we can train the association in the detection-trajectory pairs. Similarly, in the video clip we draw, we have ground truth trajectories as $\mathcal{T} = \{\tau_1, ..., \tau_k\}$. Then, for each frame $t$, we would remove the footage on and after this frame from these trajectories. It results in a new set on each frame as $\mathcal{T}^{t} = \{\emptyset\} \bigcup \{\tau^t_1, ..., \tau^t_{k^t}\}$ where $\emptyset$ is an empty trajectory. At the same time, we note the detections on the frame $t$ as $\mathcal{D}^t = \{D^t_1,...,D^t_{m^t}\}$. We then output the detection-trajectory association matrix by the introduced cross-attention. With the ground association matrix noted as $S^t$ and the estimated association matrix from softmax as $\hat{S}^t$. The loss is formulated by logistic as 
\begin{equation}
    l_{det\_traj} (\mathcal{D}, \mathcal{T}) = - \sum_{t=1}^N \sum_{\tau_i^t \in \mathcal{T}^t} \sum_{j=1}^{m^t} S^t(D^t_j, \tau^t_i) \text{log}(\hat{S}^t(D^t_j, \tau^t_i)),
\end{equation}
where an object can also be associated with an ``empty trajectory'' which means it has no corresponding existence on other frames. Finally the overall association loss is the combination of these two terms as $l_{asso} = l_{clip} + l_{det\_traj}$. For the localization stage, we can use a pretrained detection or segmentation model and freeze it or train it at the same time as training the association module. 

\paragraph{Inference.} During inference, we use an 1-stride sliding window to move from the first video clip of length $T$ to the last. In the first clip, we use the association of detections to initialize trajectories. Then, for the following steps, we do detection-trajectory and detection-detection associations at the same time. Then we use their average likelihood of association to determine the final association matrix between new-coming detections and existing trajectories. Because only one frame is new at each step of the window sliding, it is averaging the score of associating detections on the $T$-th frame and previous $T-1$ frames. The later ones have been assigned to a trajectory already. If the average association score is lower than 0.3, we start a new trajectory from the detection. In this process, we use the Hungarians algorithm to ensure the validity of the association matrix between detections and trajectories.

\section{Experiments}
\subsection{Setup}
\paragraph{Datasets and metrics.}
We choose two MOT datasets (MOT17~\cite{milan2016mot16} and Dancetrack~\cite{sun2022dancetrack}) and a MOTS dataset (MOTS20~\cite{voigtlaender2019mots}) as the experiment platforms. For evaluation, we use HOTA~\cite{luiten2021hota} as the main metric, as it has a reasonable balance between localization and association quality and evaluates association quality at a trajectory level. We also emphasize AssA as it purely measures the video-level association quality. However, on the MOTS20 test set, the HOTA evaluation protocol is not reported. So we also take IDF1 as a secondary metric to compare the quality of the association. But we still note that IDF1 is calculated at a single-frame level and cannot accurately measure the quality of association at a video level. 

\paragraph{Implementation.} 
We use ResNet-50~\cite{he2016deep} as the backbone network and BiFPN~\cite{tan2020efficientdet} for upsampling of feature maps. We use RoIAlign~\cite{he2017mask} to extract RoI of size $7 \times 7$. For a fair comparison, we follow CenterNet~\cite{zhou2019objects} for detection and keep it as-is from the pretraining on CrowdHuman~\cite{shao2018crowdhuman}. For training, the image size is $1280 \times 1280$ and we use $T=16$ to draw video clips. We use AdamW~\cite{loshchilov2017decoupled} optimizer to finetune the association module for 12K (MOT17, MOTS20) or 20K (Dancetrack) iterations with the starting learning rate of 1e-3. For segmentation, we adopt the MaskRCNN head~\cite{he2017mask} upon detection and train the head with an additional mask-rcnn loss added to the association loss. We adopt two ``linear-ReLU'' layers to project the features in the transformer. As for the evaluation of MOTS, each pixel is allowed to be assigned to at most one object; we exclusively assign pixels to at most one object per their confidence scores on MOTS20. Our implementation is based on Detectron2~\cite{wu2019detectron2}. We also refer to mmtracking~\cite{mmtrack2020} for the implementation details.

\begin{table}[]
\scriptsize
\caption{Results on  MOTS20 test set. We include only single-model methods here.}
    \centering
    \begin{tabular}{l|r|r|r|r|r|r|r}
    \toprule
         Method & sMOTSA $\uparrow$ & IDF1$\uparrow$ 	& MOTSA	$\uparrow$& FP $\downarrow$& FN $\downarrow$ & ID Sw. $\downarrow$	&Frag $\downarrow$	\\
         \midrule
         Track R-CNN~\cite{voigtlaender2019mots} & 40.6 & 42.4 & 55.2 & 1,261 & 12,641 & 567 & 868\\
          TraDes~\cite{trades} & 50.8 & 58.7 & 65.5 & 1,474 & 9,169 & 492 & -\\ 
         TrackFormer~\cite{meinhardt2022trackformer} & 54.9 & 63.6 & - & 2,233 & \textbf{7,195} & 278 &  -\\ 
         SORTS~\cite{sorts} & 55.0 & 57.3 & 68.3 & 1,076 & 8,598 & 552 & \textbf{577} \\
        Ours  & \textbf{60.0} & \textbf{68.3} & \textbf{71.7} & \textbf{634} & 8,229 & \textbf{275} & 714\\
         \bottomrule
    \end{tabular}
    \label{tab:mots_benchmark}
\end{table}

\subsection{Benchmark Results}
On the MOTS20 test set (Table~\ref{tab:mots_benchmark}), we evaluate IDF1 as the main metric. Here we only show the results from single-model methods for fairness so some others such as ReMOTS~\cite{yang2020remots} are not listed here. Our results show that the proposed method can consistently outperform existing single-model methods. 
In addition to MOTS, we also benchmark our method on MOT benchmarks of MOT17 (Table~\ref{table:mot17}) and DanceTrack (Table~\ref{table:dancetrack}). On the MOT17 test set, among transformer-based methods, our proposed method obtains the highest HOTA and AssA scores, showing its superior association performance. Moreover, compared to GTR~\cite{gtr}, which uses the same detection network as ours but no position information during association, we could see the source of our method's outperforming is the use of spatio-temporal position encoding. On the DanceTrack test set, our method also achieves the highest HOTA and AssA scores among transformer-based methods.

\begin{table}[!htp]
\centering
\caption{Results on MOT17 test set. Best results among transformer methods are \underline{underlined}.}
\setlength{\tabcolsep}{7pt}
\scriptsize
\begin{tabular}{ l | c| ccccccc}
\toprule
Tracker & Transformer & HOTA $\uparrow$ & AssA $\uparrow$ & MOTA$\uparrow$ & IDF1 $\uparrow$ & ID Sw. $\downarrow$ & FP $\downarrow$ & FN $\downarrow$ \\
\midrule
FairMOT~\cite{zhang2021fairmot} & & 59.3 & 58.0 & 73.7 & 72.3 & 3,303 & 27,507 & 117,477\\ 
PermaTrack~\cite{permatrack} & & 55.5 & 53.1 & 73.8 & 68.9 & 3,699 & 28,998 & 115,104\\ 
TraDes~\cite{trades} & &  52.7 & 50.8 & 69.1 & 63.9 & 3,555 & 20,892 & 150,060\\ 
TubeTK~\cite{pang2020tubetk} & & 48.0 & 45.1 & 63.0 & 58.6 & 4,137 & 27,060 & 177,483\\ 
ByteTrack~\cite{zhang2021bytetrack} & & 63.1 & 62.0 & \textbf{80.3} & 77.3 & 2,196 & 25,491 & \textbf{83,721}\\ 
OC-SORT~\cite{cao2022observation} & & \textbf{63.2} & \textbf{63.4} & 78.0 & \textbf{77.5} & \textbf{1,950} & \textbf{15,129} & 107,055\\
TransTrk\cite{transtrack} & \checkmark &54.1 & 47.9 & 75.2 & 63.5 & 4,614 & 50,157 & \underline{86,442}\\
TransCenter~\cite{xu2021transcenter} & \checkmark & 54.5 & 49.7 & 73.2 & 62.2 & 3,663 & \underline{23,112} & 123,738\\ 
TrackFormer~\cite{meinhardt2022trackformer} & \checkmark & - & - & 65.0 & 63.9 & 3,258 & 70,443 & 123,552\\
MOTR~\cite{motr} & \checkmark & - & - & 67.4 & 67.0 & \underline{1,992} & 32,355 & 149,400\\
GTR~\cite{gtr} & \checkmark & 59.1 & 61.6 & \underline{75.3} & 71.5 & 2,859 & 26,793 & 109,854 \\ 
MeMOT~\cite{cai2022memot} & \checkmark & 56.9 & 55.2 & 72.5 & 69.0 & 2,724 & 37,221 & 115,248\\
Ours & \checkmark & \underline{60.1} & \underline{62.1} & 75.2 & \underline{72.3} & 2,729 & 24,227 & 109,912\\
\bottomrule
\end{tabular}
\label{table:mot17}
\vspace{-2mm}
\end{table}

\begin{table}[!htp]
\centering
\caption{Results on DanceTrack test set. Best transformer-based results are \underline{underlined}.}
\setlength{\tabcolsep}{7pt}
\scriptsize
\begin{tabular}{ l | c|ccccc}
\toprule
Tracker & Transformer & HOTA $\uparrow$ & DetA $\uparrow$ & AssA $\uparrow$ & MOTA$\uparrow$ & IDF1 $\uparrow$\\
\midrule
CenterTrack~\cite{zhou2020tracking} & &41.8 & 78.1 & 22.6 & 86.8 & 35.7 \\
FairMOT~\cite{zhang2021fairmot} & &39.7 & 66.7 & 23.8 & 82.2 & 40.8\\
SORT~\cite{bewley2016simple} + YOLOX~\cite{ge2021yolox}  & &47.9 & 72.0 & 31.2 & \textbf{91.8} & 50.8 \\
DeepSORT~\cite{Wojke2018deep} + YOLOX~\cite{ge2021yolox} & &45.6 & 71.0 & 29.7 & 87.8 & 47.9\\
ByteTrack~\cite{zhang2021bytetrack} + YOLOX~\cite{ge2021yolox} & & 47.3 & 71.6 & 31.4 & 89.5 & 52.5\\
OC-SORT~\cite{cao2022observation} + YOLOX~\cite{ge2021yolox} & & \textbf{55.1} & \textbf{80.3} & \textbf{38.0} & 89.4 & \textbf{54.2}\\
TransTrk\cite{transtrack} & \checkmark&45.5 & \underline{75.9} & 27.5 & \underline{88.4} & 45.2\\
MOTR~\cite{motr} &\checkmark& 48.4 & 71.8 & 32.7 & 79.2 & 46.1\\
GTR~\cite{gtr} &\checkmark& 48.0 & 72.5 & 31.9 & 84.7 & 50.3 \\ 
Ours & \checkmark & \underline{51.9}  & 72.3 & \underline{34.6} & 84.9 & \underline{51.0}\\
\bottomrule
\end{tabular}
\label{table:dancetrack}
\end{table}

Our results on diverse datasets have shown the effectiveness of our proposed method compared to other transformer-based methods. We believe that emphasizing position information during attention and association allows the DST position encoding to outperform other methods. We will continue to further prove this through an ablation study.

\subsection{Ablation Study}
Some design choices may contribute to the performance of our proposed method. To fully validate these choices, we need segmentation annotation, but the MOTS20 evaluation server has strict access restrictions, so we have to follow the common practice~\cite{zhou2020tracking} on MOT17~\cite{milan2016mot16} to split each video in MOTS20 with the first half for training and the later half for validation in the ablation study.

To have a deeper understanding of the proposed method, the first to come is the role of DST position encoding. To verify its effectiveness, we compare it with the same architecture but without positional encoding or using classic vector positional encoding~\cite{vaswani2017attention} in Table~\ref{tab:pe_ablation}. The results clearly suggest the effectiveness of our proposed DST position encoding. Moreover, the classic positional encoding hurts the association performance, which is aligned with the observations by~\citet{gtr}. 

Furthermore, we compare the performance with and without the attention mask from segmentation on MOTS20-val. The results are reported in Table~\ref{tab:mask_ablation}. It also shows the clear advantage of using such a mask when gathering and processing the features. It agrees with the intuition that such a mask eliminates the noise from the background and potential secondary subjects in bounding boxes from the representation features.

\begin{table}[!htp]
\scriptsize
\caption{The ablation study of \textbf{positional encoding} on MOTS20-val.}
    \centering
    \begin{tabular}{l|r|r|r|r|r|r|r}
    \toprule
         pos-encode &  HOTA $\uparrow$ & IDF1 $\uparrow$& DetA $\uparrow$	& AssA  $\uparrow$	& sMOTA$ \uparrow$ & MOTSA $\uparrow$& ID Sw.$ \downarrow$	\\
         \midrule
         w/o pos-encoding & 64.4 & 72.5 & 72.5 &	58.0 & 71.6	& 82.8 & 150\\
         classic pos-encoding~\cite{vaswani2017attention} & 64.1 & 72.5 & 69.7 & 59.3 & 67.8 & 79.6 & 162 \\
         DST pos-encoding & 67.1 & 74.9	&72.8	&62.3 & 71.7 &	83.0 & 135 \\
         \bottomrule
    \end{tabular}
    \label{tab:pe_ablation}
\end{table}

\begin{table}[!htp]
\scriptsize
\caption{The ablation study of \textbf{attention mask} on MOTS20-val.}
    \centering
    \begin{tabular}{l|r|r|r|r|r|r|r}
    \toprule
          & HOTA $\uparrow$ & IDF1 $\uparrow$ & DetA $\uparrow$	& AssA $\uparrow$	& sMOTA $\uparrow$ & MOTSA $\uparrow$ & ID Sw. $\downarrow$	\\
         \midrule
         w/o mask & 64.6 & 71.3 & 72.5 & 58.1 & 71.3 & 82.6 & 156\\
         w/ mask & 67.1 & 74.9	&72.8	&62.3 & 71.7 &	83.0 & 135\\
         \bottomrule
    \end{tabular}
    \label{tab:mask_ablation}
\end{table}

The ablation studies demonstrate the effectiveness of the proposed DST position encoding as the main contribution of this work. Also, the attention mask to more accurately conclude the representation of objects is proven useful when necessary mask information is given. We note that without a segmentation mask, we can use a pretrained segmentation model or a saliency detection model to generate such masks. But this would introduce an unfair advantage, so we decide not to include it on the benchmark of MOT datasets.

\section{Conclusion}
In this work, we propose a novel dense spatio-temporal (DST) position encoding to incorporate target position information into the transformer for multi-object tracking. DST encoding leverages the property of Fourier transform to make a uniform form of position representation for both single-frame objects and trajectories across multiple frames. It shows good effectiveness in the task of multi-object tracking. While multiple previous works have failed in boosting performance with classic positional encoding, our work provides a novel and efficient paradigm for future works to do object tracking beyond just appearance matching. 

\vspace{-1em}
\section*{Acknowledgement} 
We thank the proof reading and suggestions on paper writing from Yuda Song, Erica Weng and Zhengyi Luo. This work was funded in part by NSF NRI (202417) and Department of Homeland Security (2017-DN-077-ER0001).

\bibliography{egbib}
\end{document}